\documentclass[10pt,twocolumn,letterpaper]{article}

\usepackage{cvpr}
\usepackage{times}
\usepackage{epsfig}
\usepackage{graphicx}
\usepackage{amsmath}
\usepackage{amssymb}
\usepackage{pifont}
\usepackage{color}
\usepackage{multirow}
\usepackage{subcaption}
\usepackage{times}
\usepackage{epsfig}
\usepackage{algorithm}
\usepackage[noend]{algpseudocode}
\usepackage{booktabs}
\usepackage{breakcites}
\usepackage{blindtext}
\usepackage{capt-of}
\usepackage{varwidth}
\usepackage{tabularx}

\newcolumntype{x}{>\small c}

\usepackage[T1]{fontenc}
\usepackage[utf8]{inputenc}
\usepackage{authblk}

\usepackage[pagebackref=true,breaklinks=true,letterpaper=true,colorlinks,bookmarks=false]{hyperref}

\cvprfinalcopy 

\def\cvprPaperID{1758} 

\newcommand{\datasetname}{\mbox{Outside Knowledge VQA (OK-VQA)}}
\newcommand{\abvdataset}{\mbox{OK-VQA}}

\begin{document}

\title{OK-VQA: A Visual Question Answering Benchmark Requiring \\External Knowledge}
\author[1]{Kenneth Marino\thanks{Work done during internship at Allen Institute for AI}}
\author[2]{Mohammad Rastegari}
\author[2,3]{Ali Farhadi}
\author[2]{Roozbeh Mottaghi}
\affil[1]{Carnegie Mellon University}
\affil[2]{PRIOR @ Allen Institute for AI}
\affil[3]{University of Washington}
\renewcommand\Authands{ and }
\maketitle

\begin{abstract}
Visual Question Answering (VQA) in its ideal form lets us study reasoning in the joint space of vision and language and serves as a proxy for the AI task of scene understanding. However, most VQA benchmarks to date are focused on questions such as simple counting, visual attributes, and object detection that do not require reasoning or knowledge beyond what is in the image. In this paper, we address the task of knowledge-based visual question answering and provide a benchmark, called OK-VQA, where the image content is not sufficient to answer the questions, encouraging methods that rely on external knowledge resources. Our new dataset includes more than 14,000 questions that require external knowledge to answer. We show that the performance of the state-of-the-art VQA models degrades drastically in this new setting. Our analysis shows that our knowledge-based VQA task is diverse, difficult, and large compared to previous knowledge-based VQA datasets. We hope that this dataset enables researchers to open up new avenues for research in this domain. See \url{http://okvqa.allenai.org} to download and browse the dataset.
\end{abstract}

\section{Introduction}
\begin{figure}[t]
\begin{center}
   \includegraphics[width=1\linewidth]{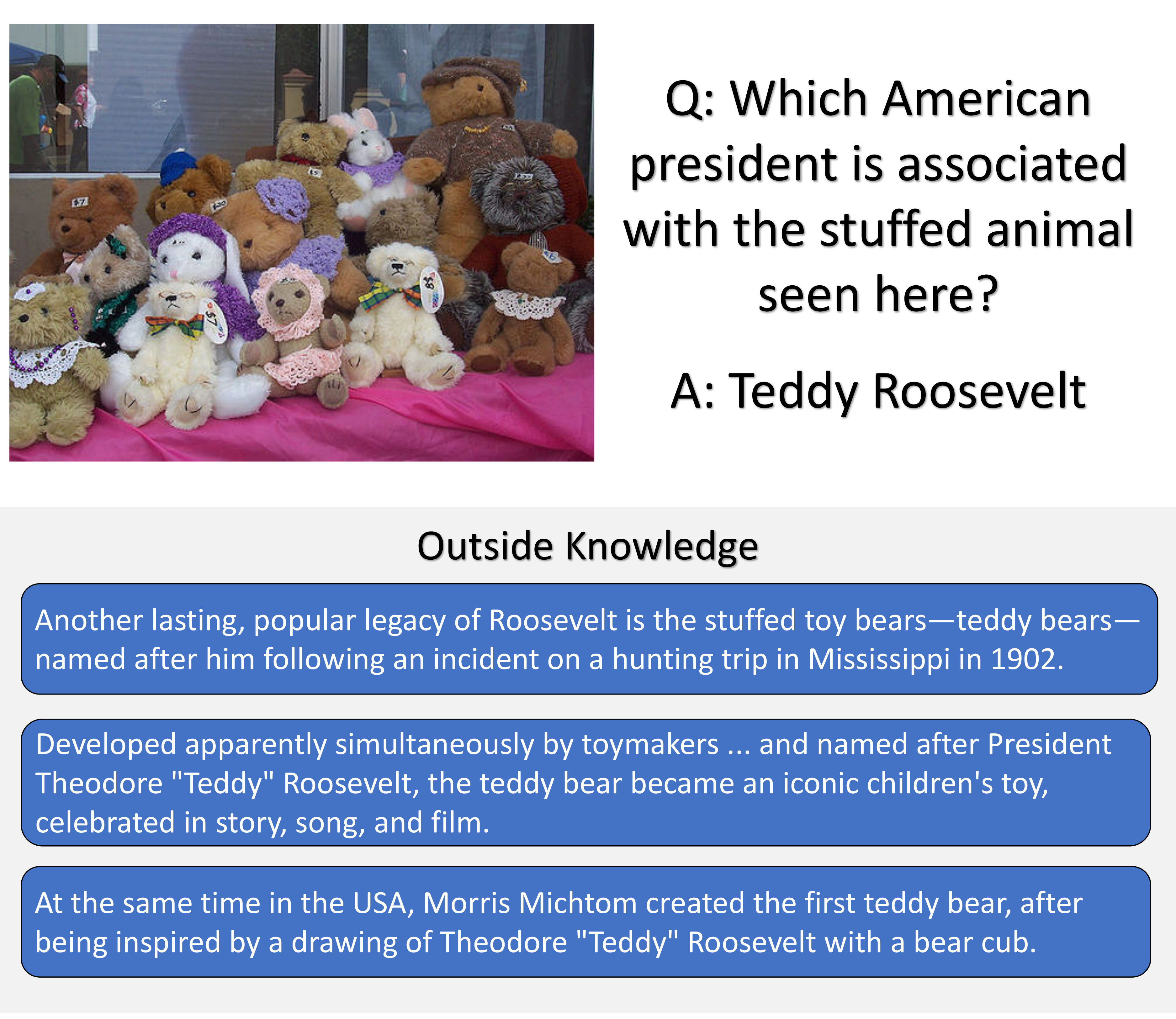}
\end{center}
\vspace{-.3cm}
   \caption{We propose a novel dataset for visual question answering, where the questions require external knowledge resources to be answered. In this example, the visual content of the image is not sufficient to answer the question. A set of facts about teddy bears makes the connection between teddy bear and the American president, which enables answering the question.}
\label{fig:teaser}
\vspace{-.5cm}
\end{figure}

The field of Visual Question Answering (VQA) has made amazing strides in recent years, achieving record numbers on standard VQA datasets~\cite{Kim2018, ben2017mutan, fukui16, jiang2018pythia}. As originally conceived, VQA is not only a fertile ground for vision and language research, but is also a proxy to evaluate AI models for the task of open-ended scene understanding. In its ideal form, VQA would require not only visual recognition, but also logical reasoning and incorporating knowledge about the world. However, current VQA datasets (e.g., \cite{antol15,zhu16})  are focused mainly on recognition, and most questions are about simple counting, colors, and other visual detection tasks, so do not require much logical reasoning or association with external knowledge. The most difficult and interesting questions ideally require knowing more than what the question entails or what information is contained in the images.

Consider the question in Figure~\ref{fig:teaser}, which asks about the relation between the teddy bear and an American president. The information in the image here is not complete for answering the question. We need to link the image content to external knowledge sources, such as the sentences at the bottom of the figure taken from Wikipedia. Given the question, image, and Wikipedia sentences, there is now enough information to answer the question: Teddy Roosevelt!

More recent research has started to look at how to incorporate knowledge-based methods into VQA~\cite{narasimhan2018out, narasimhan18, wang17a, wang17b}. These methods have investigated incorporating knowledge bases and retrieval methods into VQA datasets with a set of associated facts for each question. In this work, we go one step forward and design a VQA dataset which requires VQA to perform reasoning using unstructured knowledge.

To enable research in this exciting direction, we introduce a novel dataset, named Outside Knowledge VQA (OK-VQA), which includes only questions that require external resources for answering them. On our dataset, we can start to evaluate the reasoning capabilities of models in scenarios where the answer cannot be obtained by only looking at the image. Answering \abvdataset~questions is a challenging task since, in addition to understanding the question and the image, the model needs to: (1) learn what knowledge is necessary to answer the questions, (2) determine what query to do to retrieve the necessary knowledge from an outside source of knowledge, and (3) incorporate the knowledge from its original representation to answer the question.

The \abvdataset \ dataset consists of more than 14,000 questions that cover a variety of knowledge categories such as science \& technology, history, and sports. We provide category breakdowns of our dataset, as well as other relevant statistics to examine its properties. We also analyze state-of-the-art models and show their performance degrades on this new dataset. Furthermore, we provide results for a set of baseline approaches that are based on simple knowledge retrieval. Our dataset is diverse, difficult, and to date the largest VQA dataset focused on knowledge-based VQA in natural images.

Our contributions are: (a) we introduce the OK-VQA dataset, which includes only questions that require external resources to answer; (b) we benchmark some state-of-the-art VQA models on our new dataset and show the performance of these models degrades drastically; (c) we propose a set of baselines that exploit unstructured knowledge.

\begin{figure*}[t]
\begin{center}
   \includegraphics[width=1\linewidth]{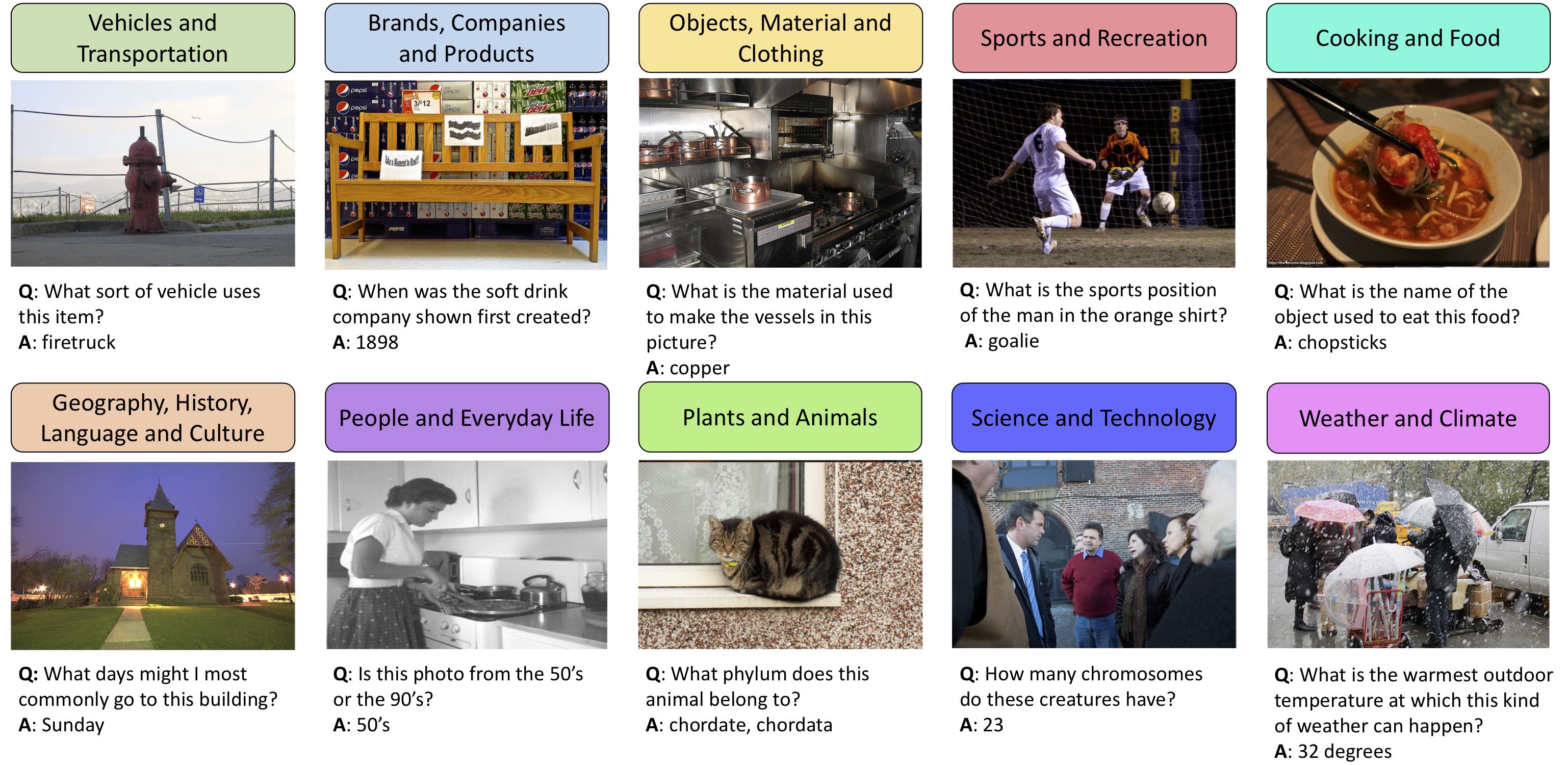}
\end{center}
\vspace{-.5cm}
   \caption{\textbf{Dataset examples.} Some example questions and their corresponding images and answers have been shown. We show one example question for each knowledge category.}
\label{fig:ckvqa}
\vspace{-.3cm}
\end{figure*}

\section{Related Work}
\noindent \textbf{Visual Question Answering (VQA).} Visual question answering (VQA) has been one of the most popular topics in the computer vision community over the past few years. Early approaches to VQA combined recurrent networks with CNNs to integrate textual and visual data \cite{malinowski15,agrawal17}. Attention-based models \cite{fukui16,lu16,xiong16,xu16,yang16,zhu16} better guide the model in answering the questions by highlighting image regions that are relevant to the question. Modular networks \cite{andreas16,hu17,johnson17b} leverage the compositional nature of the language in deep neural networks. These approaches have been extended to the video domain as well \cite{jang17,mun17,tu14}. Recently, \cite{gordon18,das18} address the problem of question answering in an interactive environment. None of these approaches, however, is designed for leveraging external knowledge so they cannot handle the cases that the image does not represent the full knowledge to answer the question.

 The problem of using external knowledge for answering questions has been tackled by \cite{wu16,wang17a,wang17b,li17,narasimhan18, narasimhan2018out}. These methods only handle the knowledge that is represented by subject-relation-object or visual concept-relation-attribute triplets, and rely on supervision to do the retrieval of facts. In contrast, answering questions in our dataset requires handling unstructured knowledge resources.

 \noindent \textbf{VQA datasets.} In the past few years several datasets have been proposed for visual question answering \cite{malinowski14b,antol15,gao15,yu15,ren15,zhu16,tapaswi16,krishna17,johnson17a,wang17b}. The DAQUAR dataset \cite{malinowski14b} includes template-based and natural questions for a set of indoor scenes. \cite{antol15} proposed the VQA dataset, which is two orders of magnitude larger than DAQUAR and includes more diverse images and less constrained answers. FM-IQA \cite{gao15} is another dataset that includes multi-lingual questions and answers. Visual Madlibs~\cite{yu15}, constructs fill-in-the-blank templates for natural language descriptions. COCO-QA \cite{ren15} is constructed automatically by converting image descriptions to questions. The idea of Visual 7W \cite{zhu16} is to provide object-level grounding for question-answer pairs as opposed to image-level associations between images and QA pairs. Visual Genome \cite{krishna17} provides dense annotations for image regions, attributes, relationships, etc. and provide free-form and region-based QA pairs for each image. MovieQA \cite{tapaswi16} is a movie-based QA dataset, where the QAs are based on information in the video clips, subtitles, scripts, etc. CLEVR \cite{johnson17a} is a synthetic VQA dataset that mainly targets visual reasoning abilities. In contrast to all these datasets, we focus on questions that cannot be answered by the information in the associated image and require external knowledge to be answered.

 Most similar to our dataset is FVQA \cite{wang17b}. While that work also tackles the difficult problem of creating a VQA dataset requiring outside knowledge, their method annotates questions by selecting a fact (a knowledge triplet such as ``dog is mammal") from a fixed knowledge base.  While this dataset is still quite useful for testing methods' ability to incorporate a knowledge base into a VQA system, our dataset tests methods' ability to retrieve relevant facts from the web, from a database, or some other source of knowledge that was not used to create the questions. Another issue is that triplets are not sufficient to represent general knowledge.

 \noindent \textbf{Building knowledge bases \& Knowledge-based reasoning.} Several knowledge bases have been created using visual data or for visual reasoning tasks \cite{zhueccv14,neil,levan,sadeghi15,zhu15,ZhuLF17}. These knowledge bases are potentially helpful resources for answering questions in our dataset. Knowledge-based question answering has received much more attention in the NLP community (e.g., \cite{berant13,yih15,yao14,bordes14,sukhbaatar15,kumar16,chen2017reading}).

\section{\abvdataset \ Dataset}
In this section we explain how we collect a dataset which better measures performance of VQA systems requiring external knowledge. The common VQA datasets such as \cite{antol15,goyal2017making} do not require much knowledge to answer a large majority of the questions. The dataset mostly contains questions such as ``How many apples are there?'', ``What animal is this?'',  and ``What color is the bowl?''. While these are perfectly reasonable tasks for open-ended visual recognition, they do not test our algorithms' ability to reason about a scene or draw on information outside of the image. Thus, for our goal of combining visual recognition with information extraction from sources outside the image, we would not be able to evaluate knowledge-based systems as most questions do not require outside knowledge.

To see this specifically, we examine the ``age annotations'' that are provided for 10,000 questions in the VQA dataset \cite{agrawal17}. For each question and image pair, an MTurk worker was asked how old someone would need to be to answer the question. While this is not a perfect metric, it is a reasonable approximation of the difficulty of a question and how much a person would have to know to answer a question. The analysis shows that more than 78\% of the questions can be answered by people who are 10 years old or younger. This suggests that very little background knowledge is actually required to answer the vast majority of these questions.
\begin{table*}[t]
 \setlength{\tabcolsep}{3pt}
  \begin{center}
    \begin{tabularx}{\linewidth}{c|c c c c c c c}
      \hline
    & \multirow{2}{*}{\parbox{1.6cm} {\centering Number of questions}} & \multirow{2}{*}{\parbox{1.5cm} {\centering Number of images}} & \multirow{2}{*}{\parbox{1.5cm} {\centering Knowledge based?}} & \multirow{2}{*}{\parbox{2cm} {\centering Goal}} & \multirow{2}{*}{\parbox{1.5cm} {\centering Answer type}} & \multirow{2}{*}{\parbox{1.2cm} {\centering Avg. A length}} & \multirow{2}{*}{\parbox{1.2cm} {\centering Avg. Q length}} \\ \\
      \hline
      DAQUAR~\cite{malinowski14b} & 12,468 & 1,449 & \ding{56} & \fontsize{8}{12}\selectfont{visual: counts, colors, objects} & Open & 1.1 & 11.5 \\
      Visual Madlibs~\cite{yu15} & 360,001 & 10,738 & \ding{56} & \fontsize{8}{12}\selectfont{visual: scene, objects, person} & FITB/MC & 2.8 & 4.9 \\
      Visual 7W~\cite{zhu16} & 327,939 & 47,300 & \ding{56} & \fontsize{8}{12}\selectfont{visual: object-grounded questions} & MC & 2.0 & 6.9 \\
      VQA (v2)~\cite{goyal2017making} & 1.1M & 200K & \ding{56} & \fontsize{8}{12}\selectfont{visual understanding} & Open/MC & 1.2 & 6.1 \\
      MovieQA~\cite{tapaswi16} & 14,944 & 408V& \ding{56} & \fontsize{8}{12}\selectfont{text+visual story comprehension} & MC & 5.3 & 9.3 \\
      CLEVR~\cite{johnson17a} & 999,968 & 100,000 & \ding{56} & \fontsize{8}{12}\selectfont{logical reasoning} & Open & 1.0 & 18.4 \\
      \hline
      KB-VQA \cite{wang17a} & 2,402 & 700 & \ding{51} & \fontsize{8}{12}\selectfont{visual reasoning with given KB} & Open & 2.0 & 6.8 \\
      FVQA~\cite{wang17b} & 5,826 & 2,190 & \ding{51} & \fontsize{8}{12}\selectfont{visual reasoning with given KB} & Open & 1.2 & 9.5 \\
      OK-VQA (ours) & 14,055 & 14,031 & \ding{51} & \fontsize{8}{12}\selectfont{visual reasoning with open knowledge} & Open & 1.3 & 8.1\\
      \hline
    \end{tabularx}
    \caption{\textbf{Comparison of various visual QA datasets.} We compare OK-VQA with some other VQA datasets. The bottom three rows correspond to knowledge-based VQA datasets. A length: answer length; Q length: question length; MC: multiple choice; FITB: fill in the blanks; KB: knowledge base.}
    \label{tab:comparison}
  \end{center}
\vspace{-.65cm}
\end{table*}

Given that current VQA datasets do not test exactly what we are looking for, we collect a new dataset. We use random images from the COCO dataset \cite{LinMBHPRDZ14}, using the original 80k-40k training and validation splits for our train and test splits. The visual complexity of these images compared to other datasets make them ideal for labeling knowledge-based questions.

In the first round of labeling, we asked MTurk workers to write a question given an image. Similar to \cite{antol15}, we prompt users to come up with questions to fool a ``smart robot.'' We also ask in the instructions that the question should be related to the image content. In addition, we prompt users not to ask what is in an image, or how many of something there is, and specify that the question should require some outside knowledge. In a second round of labeling, we asked $5$ different MTurk workers to label each question-image pair with an answer.

Although this prompt yielded many high-quality questions, it also yielded a lot of low quality questions, for example, ones that asked basic questions such as counting, did not require looking at the image, or were nonsensical. To ensure that the dataset asked these difficult knowledge-requiring questions, the MTurk provided questions were manually filtered to get only questions requiring knowledge. From a pool of 86,700 questions, we filtered down to 34,921 questions.

One more factor to consider was the potential bias in the dataset. As discussed in many works, including \cite{goyal2017making}, the VQAv1 dataset had a lot of bias. Famously, questions beginning with ``Is there a ...'' had a very strong bias towards ``Yes.'' Similarly, in our unfiltered dataset, there were a lot of questions with a bias towards certain answers. For instance, in a lot of images where there is snowfall, the question would ask ``What season is it?'' Although there were other images (such as ones with deciduous trees with multi-colored leaves) with different answers, there was a clear bias towards ``winter.'' To alleviate this problem, for train and test, we removed questions so that the answer distribution was uniform; specifically, we removed questions if there were more than 5 instances of that answer as the most common answer. This had the effect of removing a lot of the answer bias. It also had the effect of making the dataset harder by limiting the number of times VQA algorithms would see questions with a particular answer, making outside information more important. We also removed questions which had no inter-annotator agreement on the answer. Performing this filtering brought us down to 9,009 questions in train and 5,046 questions in test for a total of 14,055 questions.

Figure~\ref{fig:ckvqa} shows some of the collected questions, images, and answers from our dataset. More are provided in the Appendix~\ref{appendix:dataset_examples}. You can see that these questions require at least one piece of background knowledge to answer. For instance, in the bottom left question, the system needs to recognize that the image is of a christian church and know that those churches hold religious services on Sundays. That latter piece of knowledge should be obtained from external knowledge resources, and it cannot be inferred from the image and question alone.

\section{Dataset Statistics}
In this section, we explore the statistical properties of our dataset, and compare to other visual question answering datasets to show that our dataset is diverse, difficult, and, to the best of our knowledge, the largest VQA dataset specifically targeted for knowledge-based VQA on natural scenes.

\begin{figure}[tp]
\begin{center}
   \includegraphics[width=1\linewidth]{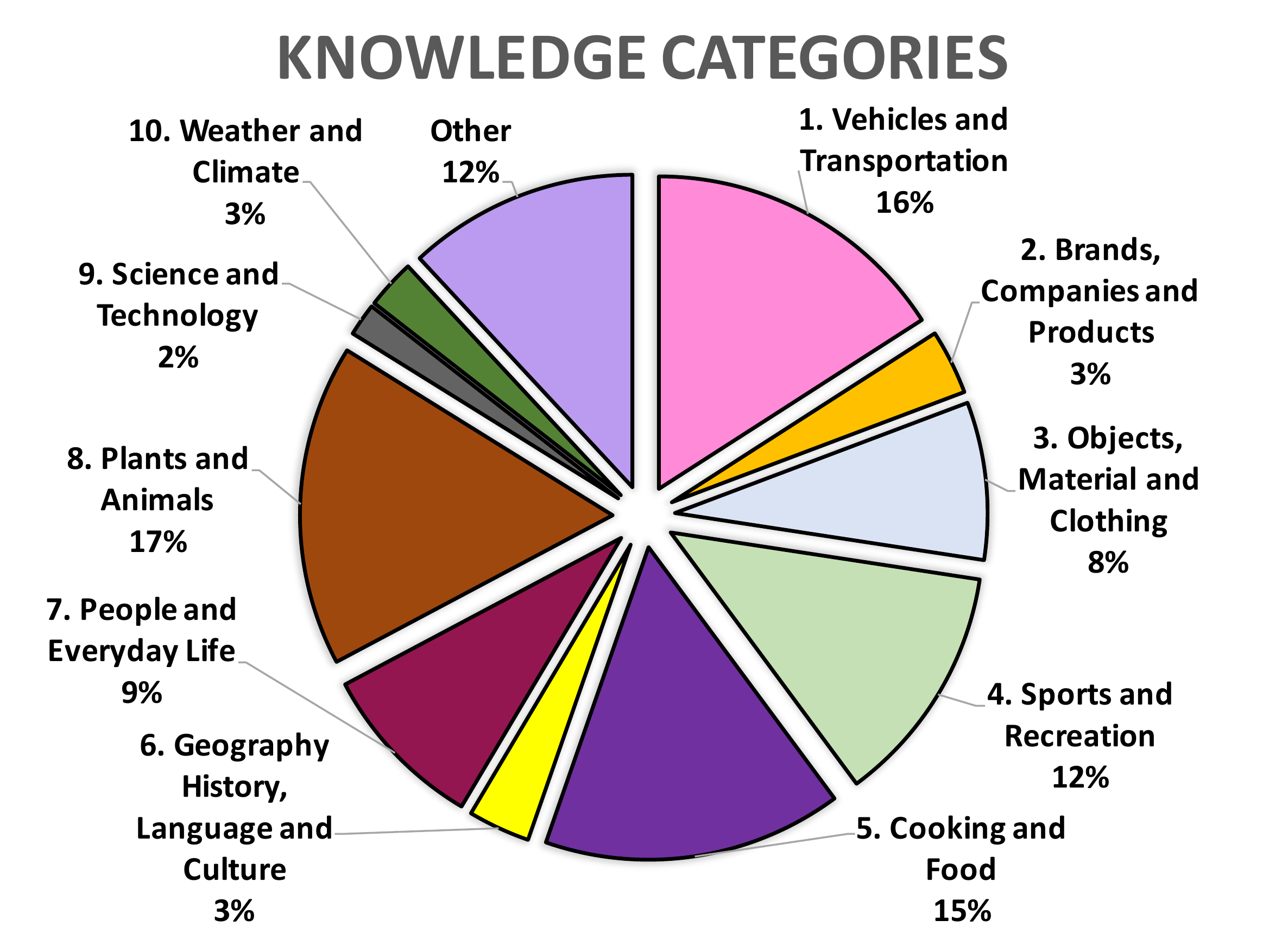}
\end{center}
   \caption{\textbf{Breakdown of questions in terms of knowledge categories.} We show the percentage of questions falling into each of our 10 knowledge categories.}
\vspace{-.3cm}
\label{fig:knowledgecats}
\end{figure}

\noindent\textbf{Knowledge category.}
Requiring knowledge for VQA is a good start, but there are many different types of knowledge that humans have about the world that could come into play. There is common-sense knowledge: water is wet, couches are found in living rooms. There is geographical knowledge: the Eiffel Tower is in Paris, scientific knowledge: humans have 23 chromosomes, and historical knowledge: George Washington is the first U.S. president. To get a better understanding of the kinds of knowledge our dataset requires, we asked five MTurk workers to annotate each question as belonging to one of ten categories of knowledge that we specified: Vehicles and Transportation; Brands, Companies and Products; Objects, Materials and Clothing; Sports and Recreation; Cooking and Food; Geography, History, Language and Culture; People and Everyday Life, Plants and Animals; Science and Technology; and Weather and Climate. If no one category had a plurality of workers, it was categorized as ``Other". This also ensured that the final category labels are mutually exclusive. We show the distribution of questions across categories in Figure~\ref{fig:knowledgecats}.

\noindent\textbf{Comparison with other VQA datasets.}
In Table~\ref{tab:comparison} we look at a number of other visual question answering datasets and compare them to our dataset in a number of different ways. In the top section, we look at a number of datasets which do not explicitly try to include a knowledge component including the ubiquitous VQAv2 dataset~\cite{goyal2017making}, the first version of which was one of the first datasets to investigate visual question answering. Compared to these datasets, we have a comparable number of questions to DAQUAR~\cite{malinowski14b} as well as MovieQA~\cite{tapaswi16}, and many more questions than knowledge-based datasets  KB-VQA~\cite{wang17a} and FVQA~\cite{wang17b}. We have fewer questions compared to CLEVR~\cite{johnson17a} where the images, questions and answers are automatically generated, as well compared to more large-scale human annotated visual datasets such as VQAv2~\cite{goyal2017making}, and Visual Madlibs~\cite{yu15}. Since we manually filtered our dataset to avoid the pitfalls of other datasets and to ensure our questions are knowledge-based and because we filtered down common answers to emphasize the long tail of answers, our dataset is more time-intensive and expensive to collect. We trade off size in this case for knowledge and difficulty.

We can see from the average question lengths and average answer lengths that our questions and answers are about comparable to  KB-VQA~\cite{wang17a} and FVQA~\cite{wang17b} and longer than the other VQA datasets with the exception of DAQUAR and CLEVR (which are partially and fully automated from templates respectively). This makes sense since we would expect knowledge-based questions to be longer as they are typically not able to be as short as common questions in other datasets such as ``How many {objects} are in the image?'' or ``What color is the couch?''.

\begin{table*}[t]
\begin{center}
\small\begin{tabularx}{\linewidth}{c|c|X*{11}{c}}
\toprule
Method & OK-VQA & VT & BCP & OMC & SR & CF & GHLC & PEL & PA & ST & WC & Other \\ \midrule
Q-Only & 14.93 & 14.64 & 14.19 & 11.78 & 15.94 & 16.92 & 11.91 & 14.02 & 14.28 & 19.76 & 25.74 & 13.51 \\
MLP & 20.67 & 21.33 & 15.81 & 17.76 & 24.69 & 21.81 & 11.91 & 17.15 & 21.33 & 19.29 & 29.92 & 19.81 \\
ArticleNet (AN) & 5.28 & 4.48 & 0.93 & 5.09 & 5.11 & 5.69 & 6.24 & 3.13 & 6.95 & 5.00 & 9.92 & 5.33 \\
BAN \cite{Kim2018} & 25.17 & 23.79 & 17.67 & 22.43 & 30.58 & 27.90 & \textbf{25.96} & 20.33 & 25.60 & 20.95 & \textbf{40.16} & 22.46 \\
MUTAN \cite{ben2017mutan} & 26.41 & 25.36 & 18.95 & 24.02 & 33.23 & 27.73 & 17.59 & 20.09 & \textbf{30.44} & 20.48 & 39.38 & 22.46 \\
BAN + AN & 25.61 & 24.45 & 19.88 & 21.59 & 30.79 & 29.12 & 20.57 & 21.54 & 26.42 & \textbf{27.14} & 38.29 & 22.16 \\
MUTAN + AN & \textbf{27.84} & \textbf{25.56} & \textbf{23.95} & \textbf{26.87} & \textbf{33.44} & \textbf{29.94} & 20.71 & \textbf{25.05} & 29.70 & 24.76 & 39.84 & \textbf{23.62}  \\
\hline
BAN/AN oracle & 27.59 & 26.35 & 18.26 & 24.35 & 33.12 & 30.46 & 28.51 & 21.54 & 28.79 & 24.52 & 41.4 & 25.07 \\
MUTAN/AN oracle & 28.47 & 27.28 & 19.53 & 25.28 & 35.13 & 30.53 & 21.56 & 21.68 & 32.16 & 24.76 & 41.4 & 24.85 \\
\bottomrule
\end{tabularx}
\end{center}
\vspace{-.3cm}
\caption{\textbf{Benchmark results on \abvdataset.} We show the results for the full OK-VQA dataset and for each knowledge category: Vehicles and Transportation (VT); Brands, Companies and Products (BCP); Objects, Material and Clothing (OMC); Sports and Recreation (SR); Cooking and Food (CF); Geography, History, Language and Culture (GHLC); People and Everyday Life (PEL); Plants and Animals (PA); Science and Technology (ST); Weather and Climate (WC); and Other.}
\vspace{-.5cm}
\label{table:VQA}
\end{table*}

\noindent\textbf{Question statistics.}
We also collected statistics for our dataset by looking at the number of questions, and by looking at which were most frequent for each knowledge category. OK-VQA has 12,591 unique questions out of 14,055 total, and 7,178 unique question words. This indicates that we get a variety of different questions and answers in our dataset.
\begin{figure}[h]
\begin{center}
   \includegraphics[width=\linewidth]{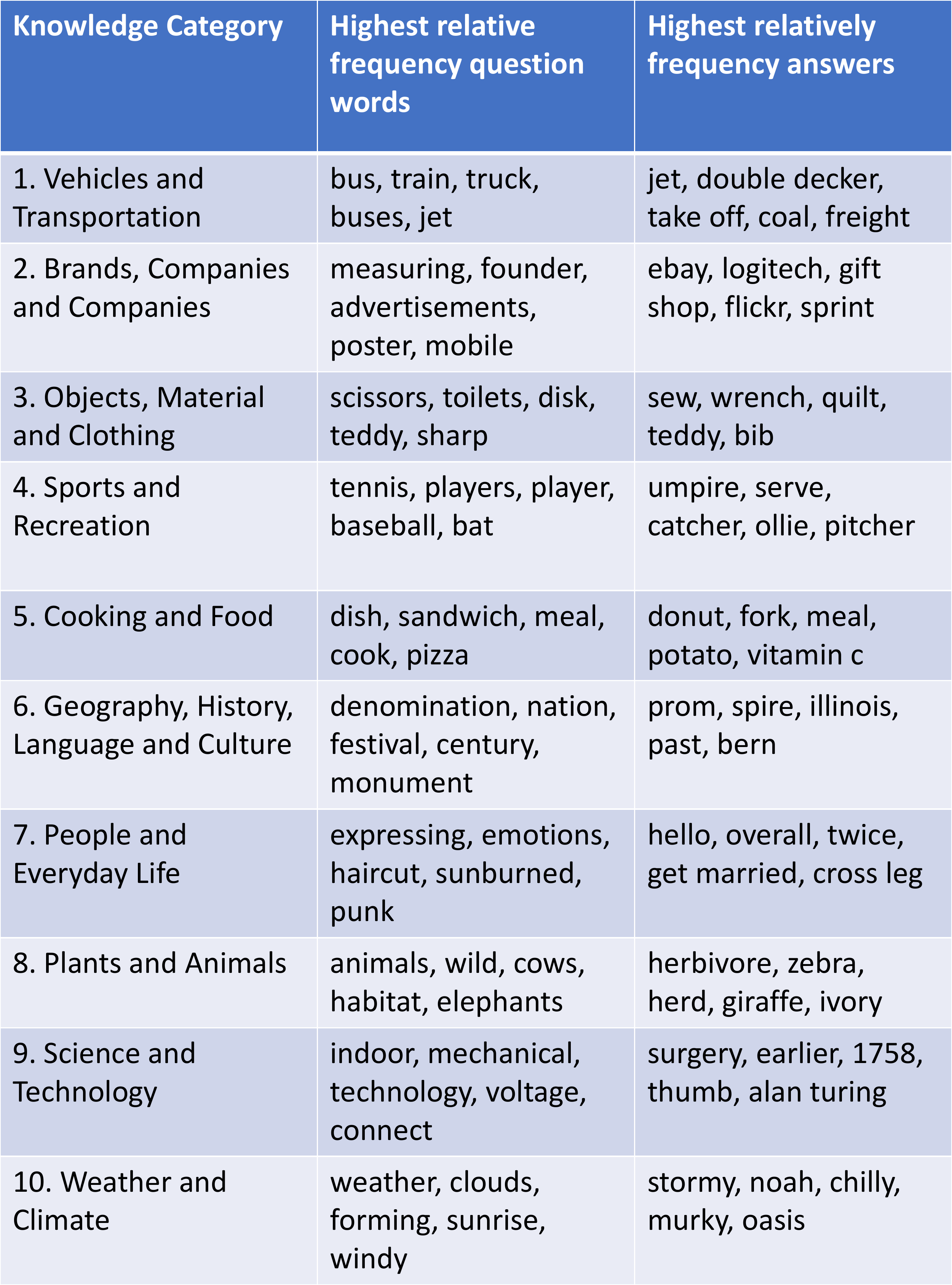}
\end{center}
\vspace{-0.5cm}
   \caption{For each category we show the question words and answers that have the highest relative frequency across our knowledge categories (i.e. frequency in category divided by overall frequency).}
\vspace{-.65cm}
\label{fig:mostcommon}
\end{figure}
We also looked at the variety of images in our dataset. As we stated earlier, our images come from the COCO image dataset, so our dataset contains the same basic distribution of images. However, we only use a subset of COCO images, so we want to see if we still get a wide distribution of images. For this, we ran a  Places2 \cite{zhou2017places} classifier on our images and looked at the top-1 scene class for each image and compared that to COCO overall. Out of 365 scenes, our dataset contains all but 5 classes: hunting lodge, mansion, movie theater, ruin and volcano. These classes appear infrequently in the overall COCO dataset (10, 22, 28, 37 and 25 times respectively), so overall, we still captured quite a lot of the variation in scenes.

Finally, we show in Figure~\ref{fig:mostcommon} the question words and answers in each category that are the most ``unique'' to get a better idea of what types of questions we have in each of these categories. We calculate these for each knowledge category by looking at the number of appearances within the category over the total number in the dataset to see which question words and answers had the highest relative frequency in their category. When looking at the question words, we see words specific to categories such as bus in Vehicles and Transportation, sandwich in Cooking and Food, and clouds in Weather and Climate. We also see that the answers are also extremely related to each category, such as herbivore in Plants and Animals, and umpire in Sports and Recreation. In Appendix~\ref{appendix:dataset_stats}, we also show the most common question words and answers.

\section{Benchmarking}

In this section, we evaluate current state-of-the-art VQA approaches and provide results for some baselines, including knowledge-based ones.

\textbf{MUTAN}~\cite{ben2017mutan}:
Multimodal Tucker Fusion (MUTAN) model \cite{ben2017mutan}, a recent state-of-the-art tensor-based method for VQA. Specifically, we use the attention version of MUTAN, and choose the parameters to match the single best performing model of \cite{ben2017mutan}.

\textbf{BAN}~\cite{Kim2018}:
Bilinear Attention Networks for VQA. A recent state-of-the art VQA method that uses a co-attention mechanism between the question features and the bottom-up detection features of the image. We modify some hyperparameters to improve performance on our dataset (see Appendix~\ref{appendix:combine_details}.

\textbf{MLP}:
The MLP has 3 hidden layers with ReLU activations and hidden size $2048$ that concatenates the image and question features after a skip-thought GRU after one fully connected layer each. Like MUTAN, it uses fc7 features from ResNet-152.

\textbf{Q-Only}:
The same model as MLP, but only takes the question features.

\textbf{ArticleNet (AN)}:
We consider a simple knowledge-based baseline that we refer to as ArticleNet. The idea is to retrieve some articles from Wikipedia for each question-image pair and then train a network to find the answer in the retrieved articles.

Retrieving articles is composed of three steps. First, we collect possible search queries for each question-image pair. We come up with all possible queries for each question by combining words from the question and words that are identified by pre-trained image and scene classifiers. Second, we use the Wikipedia search API to get the top retrieved article for each query. Third, for each query and article, we extract a small subset of each article that is most relevant for the query by selecting the sentences within the article that best correspond to our query based on the frequency of those query words in the sentence.

\begin{figure*}[t]
\begin{center}
   \includegraphics[width=0.9\linewidth]{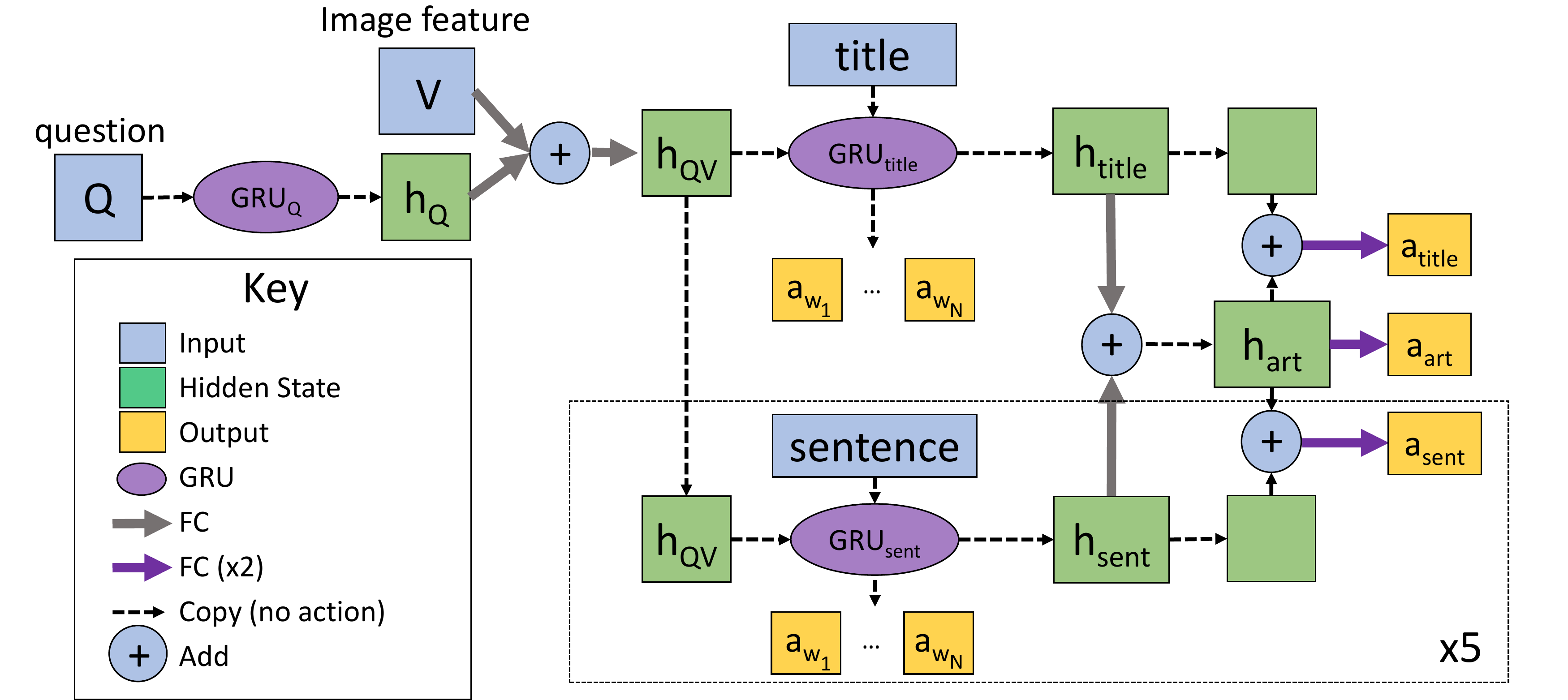}
\end{center}
   \vspace{-.2cm}
   \caption{\textbf{ArticleNet architecture.} ArticleNet takes in the question $Q$ and visual features $V$. All modules within the dotted line box share weights. The output of the GRUs is used to classify each word as the answer or not $a_{w_i}$. The final GRU hidden states $h_{title}$ and $h_{sent}$ are put through fully connected layers to predict if the answer is in the sentence $a_{sent}$ or title $a_{title}$, and then are combined together and used to classify if the answer is in the article $a_{art}$.}
\label{fig:articleNet}
\end{figure*}

Once the sentences have been retrieved, the next step is to filter and encode them for use in VQA. Specifically, we train ArticleNet to predict whether and where the ground truth answers appear in the article and in each sentence. The architecture is shown in Figure \ref{fig:articleNet}. To find the answer to a question, we pick the top scoring word among the retrieved sentences. More specifically, we take the highest value of $a_{w_i}.a_{sent}$, where $a_{w_i}$ is the score for the word being the answer and $a_{sent}$ is the score for the sentence including the answer.
For a more detailed description of ArticleNet see Appendix~\ref{appendix:articlenet}.

\textbf{MUTAN + AN}:
We augment MUTAN with the top sentence hidden states ($h_{sent}$ in Figure~\ref{fig:articleNet}) from ArticleNet (AN). During VQA training and testing, we take the top predicted sentences (ignoring duplicate sentences), and feed them in the memory of an end-to-end memory network \cite{sukhbaatar15}. The output of the memory network is concatenated with the output of the first MUTAN fusion layer.

\textbf{BAN + AN}:
Similarly, we incorporate the ArticleNet hidden states into BAN and incorporate it into VQA pipeline with another memory network. We concatenate output of the memory network with the BAN hidden state right before the final classification network. See Appendix~\ref{appendix:combine_details} for details.

\textbf{MUTAN/AN oracle}:
As an upper bound check, and to see potentially how much VQA models could benefit from the knowledge retrieved using ArticleNet, we also provide results on an oracle, which simply takes the raw ArticleNet and MUTAN predictions, taking the best answer (comparing to ground truth) from either.

\textbf{BAN/AN oracle}:
Similar to the MUTAN/AN oracle, but we take the best answer from the raw ArticleNet and BAN instead, again taking the best answer for each question.

\noindent \textbf{Benchmark results.} We report the results using the common VQA evaluation metric \cite{antol15}, but use each of our answer annotations twice, since we have 5 answer annotations versus 10 in \cite{antol15}. We also stem the answers using Porter stemming to consolidate answers that are identical except for pluralization and conjugation as in ~\cite{wang17b}. We also show the breakdowns for each of our knowledge categories. The results are reported in Table~\ref{table:VQA}.

The first observation is that no method gets close to numbers on standard VQA dataset such as VQA~\cite{goyal2017making} (where the best real open-ended result for the 2018 competition is 72.41). Moreover, state-of-the-art models such as MUTAN \cite{ben2017mutan} and BAN \cite{Kim2018}, which are specifically designed for VQA to learn high-level associations between the image and question, get far worse numbers on our dataset. This suggests that \abvdataset~cannot be solved simply by coming up with a clever model, but actually requires methods that incorporate information from outside the image.

It is interesting to note that although the performance of the raw ArticleNet is low, it provides improvement when combined with the state-of-the-art models (MUTAN + AN and BAN + AN). From the oracle numbers, we can see that the knowledge retrieved by ArticleNet provides complementary information to the state-of-the-art VQA models. These oracles are optimistic upper bounds using ArticleNet, but they show that smarter knowledge-retrieval approaches could have stronger performance on our dataset. Note that ArticleNet is not directly trained on VQA and can only predict answers within the articles it has retrieved. So the relatively low performance on VQA is not surprising. 
\begin{figure*}[t]
\begin{center}
\vspace{-.1cm}
   \includegraphics[width=\linewidth]{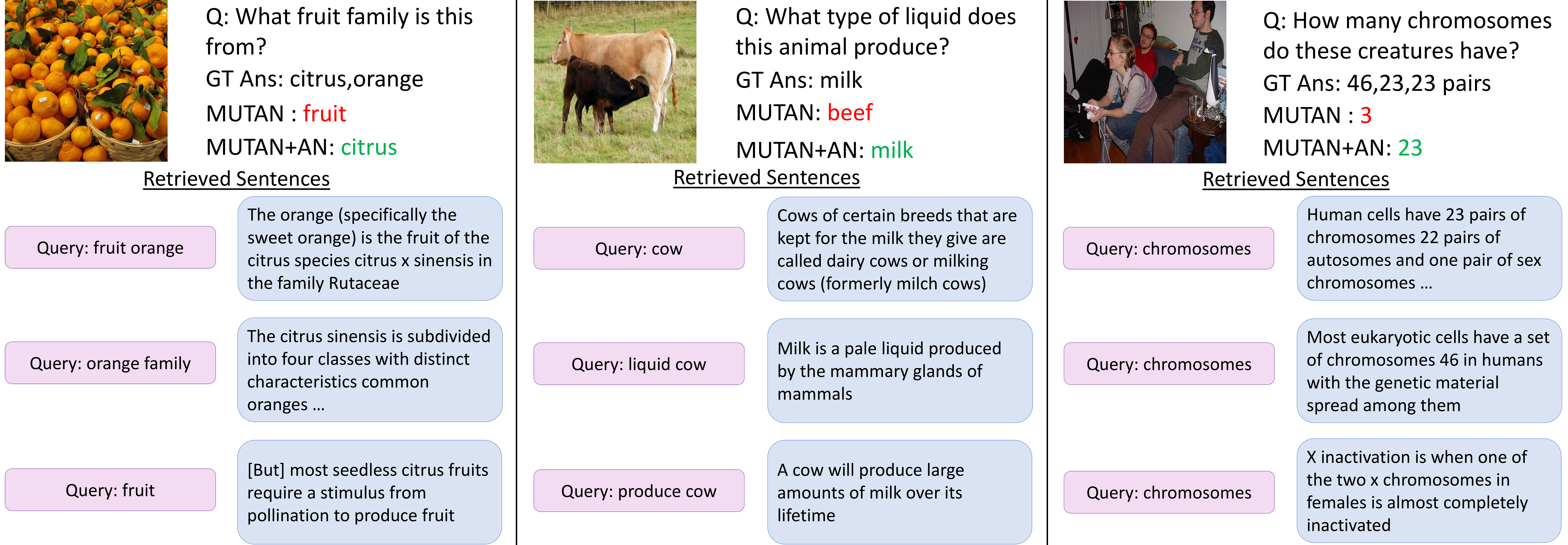}
\end{center}
   \vspace{-.2cm}
   \caption{\textbf{Qualitative results.} We show the result of MUTAN+AN compared to the MUTAN baseline answer and the ground truth answer (`GT Ans'). We show the query words that were used by ArticleNet (pink boxes) and the corresponding most relevant sentences (blue boxes).}
\label{fig:qualitative}
\vspace{-.3cm}
\end{figure*}
Looking at the category breakdowns, we see that ArticleNet is particularly helpful for brands, science, and cooking categories, perhaps suggesting that these categories are better represented in Wikipedia. It should be noted that the major portion of our dataset requires knowledge outside Wikipedia such as commonsense or visual knowledge.

The Q-Only baseline performs significantly worse than the other VQA baselines, suggesting that visual features are indeed necessary and our procedure for reducing answer bias was effective.
\begin{table}[t]
\begin{center}
\begin{tabular}{cc}
\toprule
      Method & VQA score on \abvdataset \\ \midrule
      ResNet152 & 26.41 \\
      ResNet50 & 24.74 \\
      ResNet18 & 23.64 \\
      Q-Only & 14.93 \\ \bottomrule
      \bottomrule
\end{tabular}
\end{center}
\vspace{-.5cm}
\caption{Results on \abvdataset \ with different visual features.}
\vspace{-.5cm}
\label{table:student}
\end{table}
\noindent \textbf{Visual feature ablation.} We also want to demonstrate the difficulty of the dataset from the perspective of visual features, so we show MUTAN results using different ResNet architectures. The previously reported result for MUTAN is based on ResNet152. We also show the results using extracted features from ResNet50 and ResNet18 in Table~\ref{table:student}. From this table it can be seen that going from ResNet50 to ResNet152 features only has a marginal improvement, and similarly going from ResNet18 to ResNet50. However, going from ResNet18 to no image (Q-Only) causes a large drop in performance. This suggests that our dataset is indeed visually grounded, but better image features do not hugely improve the results, suggesting the difficulty lies in the retrieving the relevant knowledge and reasoning required to answer the questions.

\begin{figure}[t]
\begin{center}
    \includegraphics[width=\linewidth]{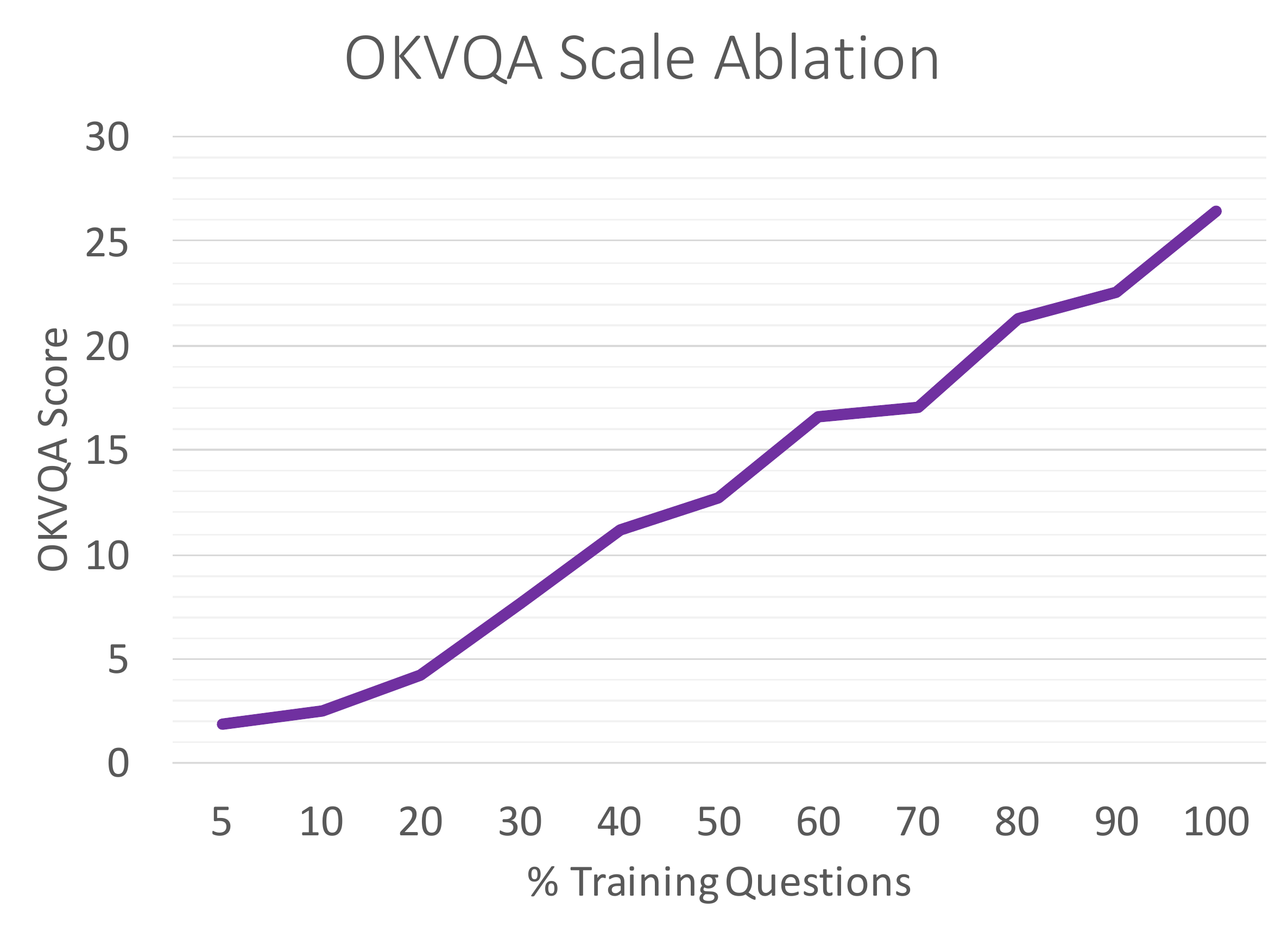}
\end{center}
\vspace{-0.5cm}
   \caption{Results on \abvdataset \ using different sizes of the training set.}
\vspace{-.3cm}
\label{fig:trainingsize}
\end{figure}

\noindent \textbf{Scale ablation.} Finally, we investigate the degree to which the size of our dataset relates to its difficulty as opposed to the nature of the questions themselves. We first take a random subdivision of our training set and train MUTAN on progressively smaller subsets of the training data and evaluate on our original test set. Figure~\ref{fig:trainingsize} shows the results.

\noindent \textbf{Qualitative examples.} We show some qualitative examples in Figure~\ref{fig:qualitative} to see how outside knowledge helps VQA systems in a few examples. We compare MUTAN+AN method with MUTAN. The left example asks what ``fruit family'' the fruit in the image (oranges) comes from. We see that two sentences that directly contain the information that oranges are citrus fruits are retrieved ---``The orange ... is a fruit of the citrus species'' and ``The citrus sinensis is subdivided into four classes [including] common oranges''.

The middle example asks what liquid the animal (cow) produces. The first and third sentences tell us that cows produce milk, and the second sentence tells us that milk is a liquid. This gives the combined MUTAN+AN method enough information to correctly answer milk.

The example on the right asks how many chromosomes humans have. It is somewhat ambiguous whether it means how many individual chromosomes or how many pairs, so workers labeled both as answers. The retrieved articles are helpful here, retrieving two different articles referring to 23 pairs of chromosomes and 46 chromosomes total. The combined MUTAN+AN method correctly answers 23, while MUTAN guesses 3.

\section{Conclusion}
We address the task of knowledge-based visual question answering. We introduce a novel benchmark called \abvdataset \ for this task. Unlike the common VQA benchmarks, the information provided in the question and the corresponding images of \abvdataset \ is not sufficient to answer the questions, and answering the questions requires reasoning on external knowledge resources. We show that the performance of state-of-the-art VQA models significantly drops on \abvdataset. We analyze the properties and statistics of the dataset and show that background knowledge can improve results on our dataset. Our experimental evaluations show that the proposed benchmark is quite challenging and that there is a large room for improvement.

\noindent {\footnotesize {\bf Acknowledgements}: We would like to thank everyone who took time to review this work and provide helpful comments. This work is in part supported by NSF IIS-165205, NSF IIS-1637479, NSF IIS-1703166, Sloan Fellowship, NVIDIA Artificial Intelligence Lab, and Allen Institute for artificial intelligence. Thanks to Aishwarya Agrawal, Gunnar Sigurdsson, Victoria Donley, Achal Dave, and Eric Kolve who provided valuable assistance, advice and feedback. Kenneth Marino is supported by the Department of Defense (DoD) through the National Defense Science \& Engineering Graduate Fellowship (NDSEG) Program. This research is sponsored by Army Research Office and accomplished under Grant Number W911NF-18-1-0019.}

{\small
\bibliographystyle{ieee}
\bibliography{egbib}
}

\clearpage
\appendix

\section{More Dataset Statistics}
\label{appendix:dataset_stats}

\begin{figure}[h]
\begin{center}
\vspace{-.1cm}
   \includegraphics[width=0.9\linewidth]{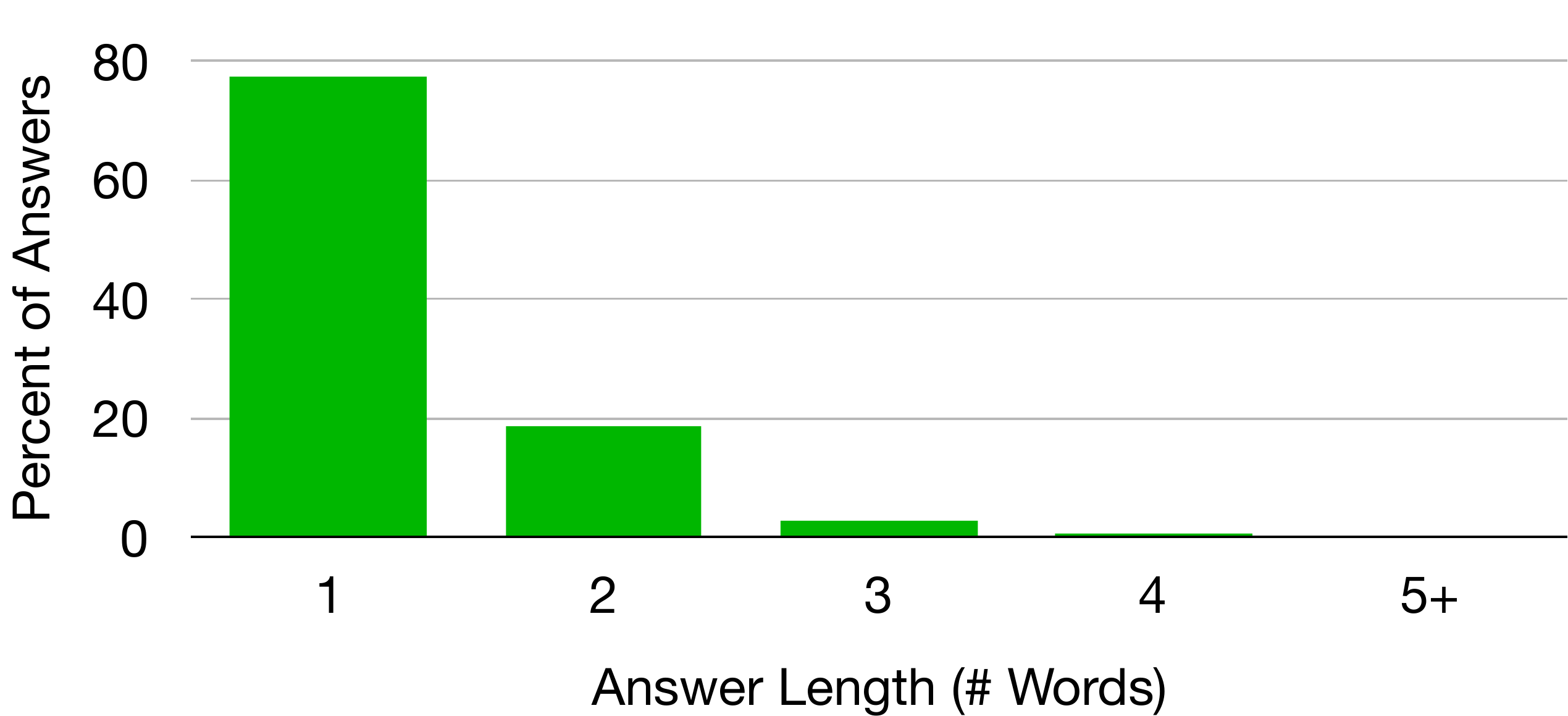}
\end{center}
\vspace{-.3cm}
   \caption{\textbf{Answer length distribution.} Histogram of the answer lengths in the dataset.}
\label{fig:anslen}
\vspace{-.2cm}
\end{figure}

\begin{figure}[h]
\begin{center}
\vspace{-.1cm}
   \includegraphics[width=\linewidth]{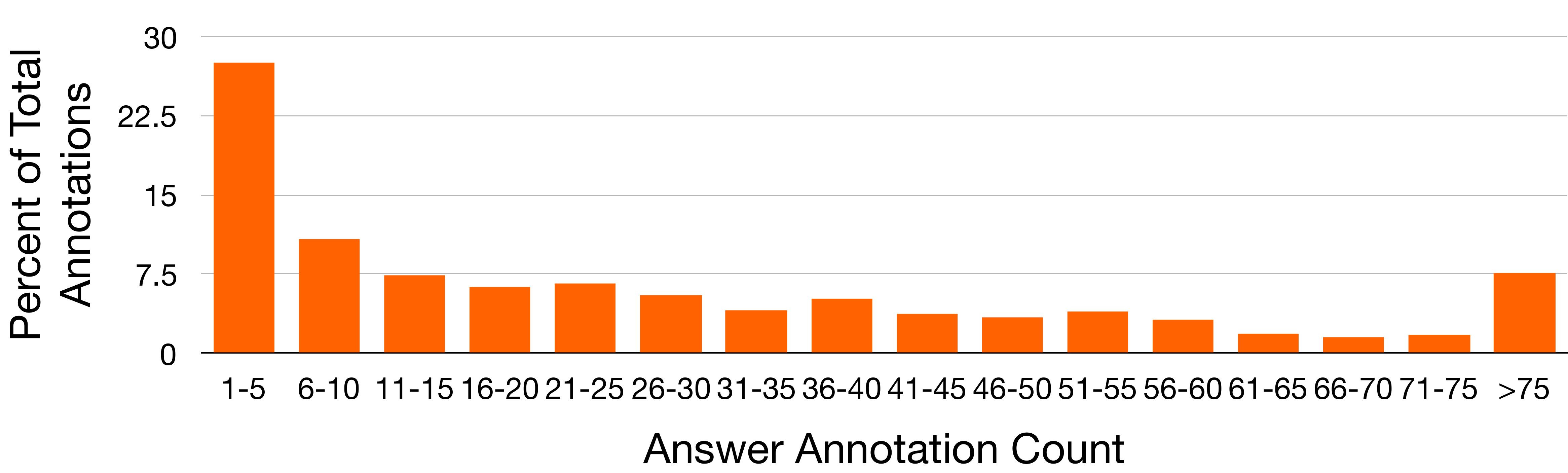}
\end{center}
   \vspace{-.3cm}
   \caption{\textbf{Answer frequency distribution.} Histogram of the frequency of answers in the dataset. All 5 answers for each question are considered to compute the histogram. This shows for instance that answers that appear between 6 and 10 times in the dataset make up about 10\% of all answers.}
\label{fig:ansfreq}
\vspace{-.2cm}
\end{figure}

In Figure~\ref{fig:anslen} we show the distribution of the lengths of the answers in the dataset. In Figure~\ref{fig:ansfreq}, we show the distribution of answer frequency for each of the unique answers in the dataset.

\begin{figure}[h]
\begin{center}
   \includegraphics[width=\linewidth]{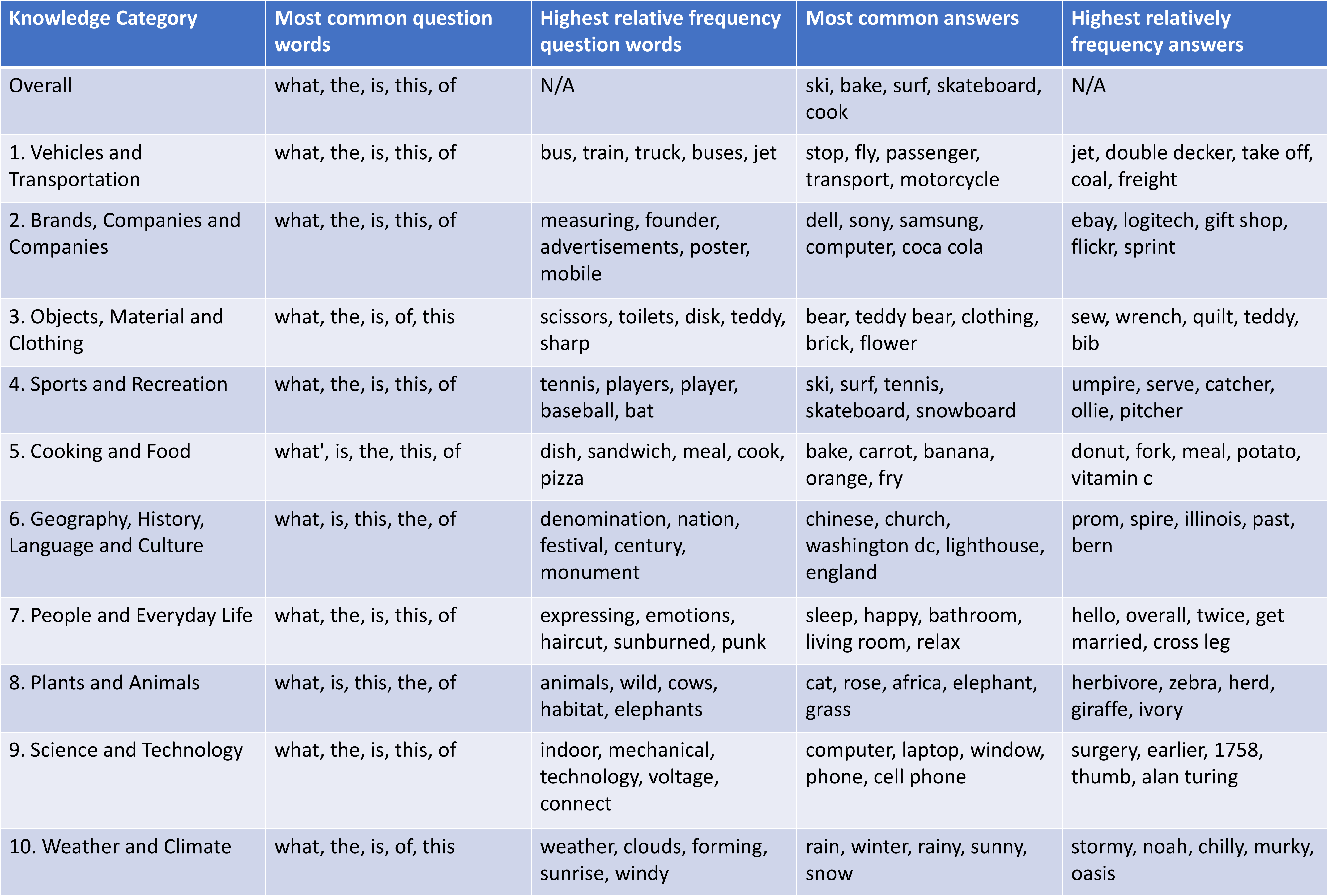}
\end{center}
   \caption{For each category we show the question words and answers that have the highest frequency and relative frequency across our knowledge categories (i.e. frequency in category divided by overall frequency).}
\label{fig:mostcommonsupp}
\end{figure}

In Figure~\ref{fig:mostcommonsupp} we show the most common and the highest ``relative frequency'' question words and answers in each category.

Some other relevant dataset statistics for \abvdataset~can be found in Table~\ref{table:randomstats}.

\begin{table}[h]
\begin{center}
\begin{tabular}{cc}
\toprule
    \midrule
      Number of unique answers & 14,454 \\
      Test set covered by top 2000 answers & 88.5057\% \\
      Number of unique questions & 12,591\\
      Number of unique question words & 7,178 \\ \bottomrule
      \bottomrule
\end{tabular}
\end{center}
\vspace{-.5cm}
\caption{More \abvdataset \ dataset statistics.}
\vspace{-.5cm}
\label{table:randomstats}
\end{table}

\clearpage

\section{ArticleNet Details} \label{ArticleNet}
\subsection{Article Collection} \label{QueryArticle}
Extracting the articles is composed of three steps: collecting possible search queries, using the Wikipedia search API to get the top retrieved article for each query and extracting a small subset of each article that is most relevant for the query.

\begin{figure*}[h]
\begin{center}
  \includegraphics[width=\linewidth]{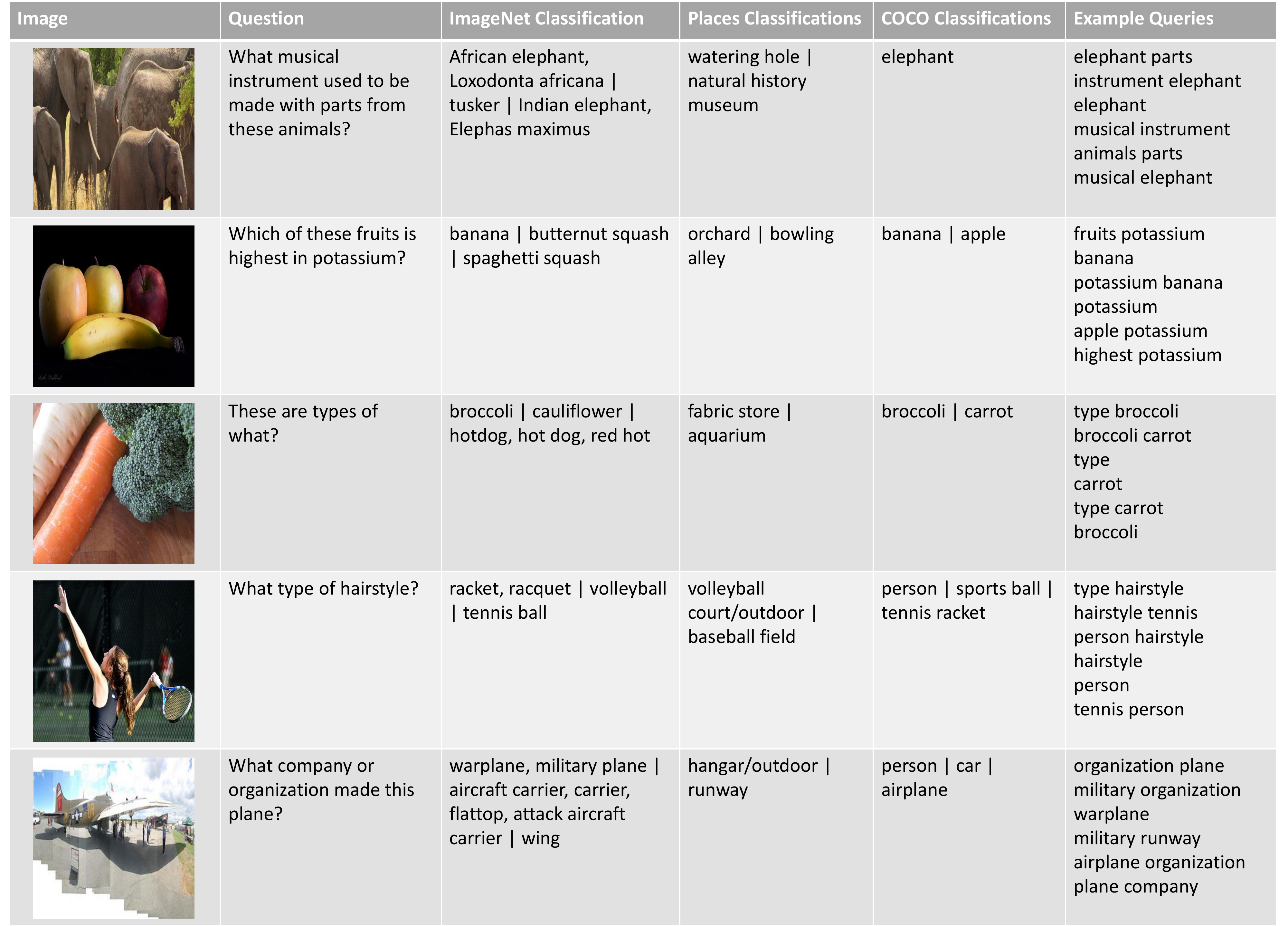}
\end{center}
  \caption{\textbf{Example generated queries.} For some example questions, we show the image, question and top classification results from the trained models. In the rightmost column, we show some example queries that can be constructed for each question.}
\label{fig:queries}
\end{figure*}

To perform the search query step, first we need to come up with possible queries for each question. We extract all non-stop words (i.e. remove ``the'', ``a'', ``what'', etc.) from the question itself. Next we extract visual entities from the images by taking the top classifications from trained classifiers. We take the top classifications from an object classifier (trained on ImageNet \cite{russakovsky2015imagenet}), a place classifier (trained on Places2 \cite{zhou2017places}) and an object detector \cite{girshick2015fast,chen2017implementation} (trained on COCO dataset \cite{LinMBHPRDZ14}). Figure~\ref{fig:queries} shows some example images and their corresponding questions and classifications and shows some example queries that can be generated for that question.

Once the query words are selected, we compute possible queries. We choose every query word by itself and every two word combination of the query words as possible queries. We then retrieve the top article for each query from the Wikipedia search. Using the retrieval model from \cite{chen2017reading} to achieve a consistent snapshot, we retrieve the raw text.

Finally, we use the original query and the retrieved Wikipedia article to extract the most relevant sentences from the article for the query. Essentially, we perform another step of retrieval. The sentence priority is determined by three hierarchical metrics: (1) the number of unique query words in the sentence, (2) the total number of query words in the sentence, counting repeats, (3) the order of the sentence in the article. The priority is determined by factor (1). If two sentences tie on this metric, we use metric (2) as a tie breaker, and similarly we use metric (3) to break ties for metric (2).

After these steps, we have our final ``article'' for each query consisting of the title, and $T$ most relevant sentences (in our case $T=5$). In our experiments, we retrieve on the order of $100$ articles for each question at this step. 

\subsection{ArticleNet Overview}
\label{appendix:articlenet}
Once the Wikipedia articles have been retrieved, the next step is to filter and encode them for use in VQA. Simple encodings such as an average word2vec encoding, or with skip-thought \cite{kiros2015skip} are not suitable for encoding long articles. Hence, we train an encoding specific to our data and useful for our eventual task. Taking inspiration from the fact that a hidden layer of a network trained on ImageNet is a good representation for images, we train a network on the retrieved articles on a proxy task to get a good representation. Specifically, we train ArticleNet to predict whether and where the ground truth answers appear in the article and each sentence. This also gives a way to narrow down the hundreds of articles for each question-image pair to a handful for the final VQA training.

For each of the Wikipedia articles, each word and series of words in the sentence are compared to the ground truth answer for that question to see if they match (using Porter stemming). Hence, a label $l_{art}$ is obtained if the answer appears in the article, and also a label $l_{title}$ and $l_{sent_i}$ if the answer appears in the title or sentence $i$, and a label $l_{word_j}$ for each word in the title and sentence.

The architecture of the ArticleNet is shown in Figure~\ref{fig:articleNet}. The inputs to the network are the question $Q$, the visual features $V$ taken from an ImageNet trained ResNet152 \cite{he2016deep}, the title of the Wikipedia article, and the $T$ sentences of the article (retrieved by the method explained in the previous section). From these inputs, it predicts whether the answer is in the title $a_{title}$, any of the sentences $a_{sent}$ or the entire article $a_{art}$. The hidden states of this network are used later in the VQA pipeline to encode the sentences.

After training, the network is evaluated on the articles for each question, the sentences that have the highest prediction score $a_{sent}$ are used in our VQA training.
\begin{figure}[H]
     \centering
     \includegraphics[width=\linewidth]{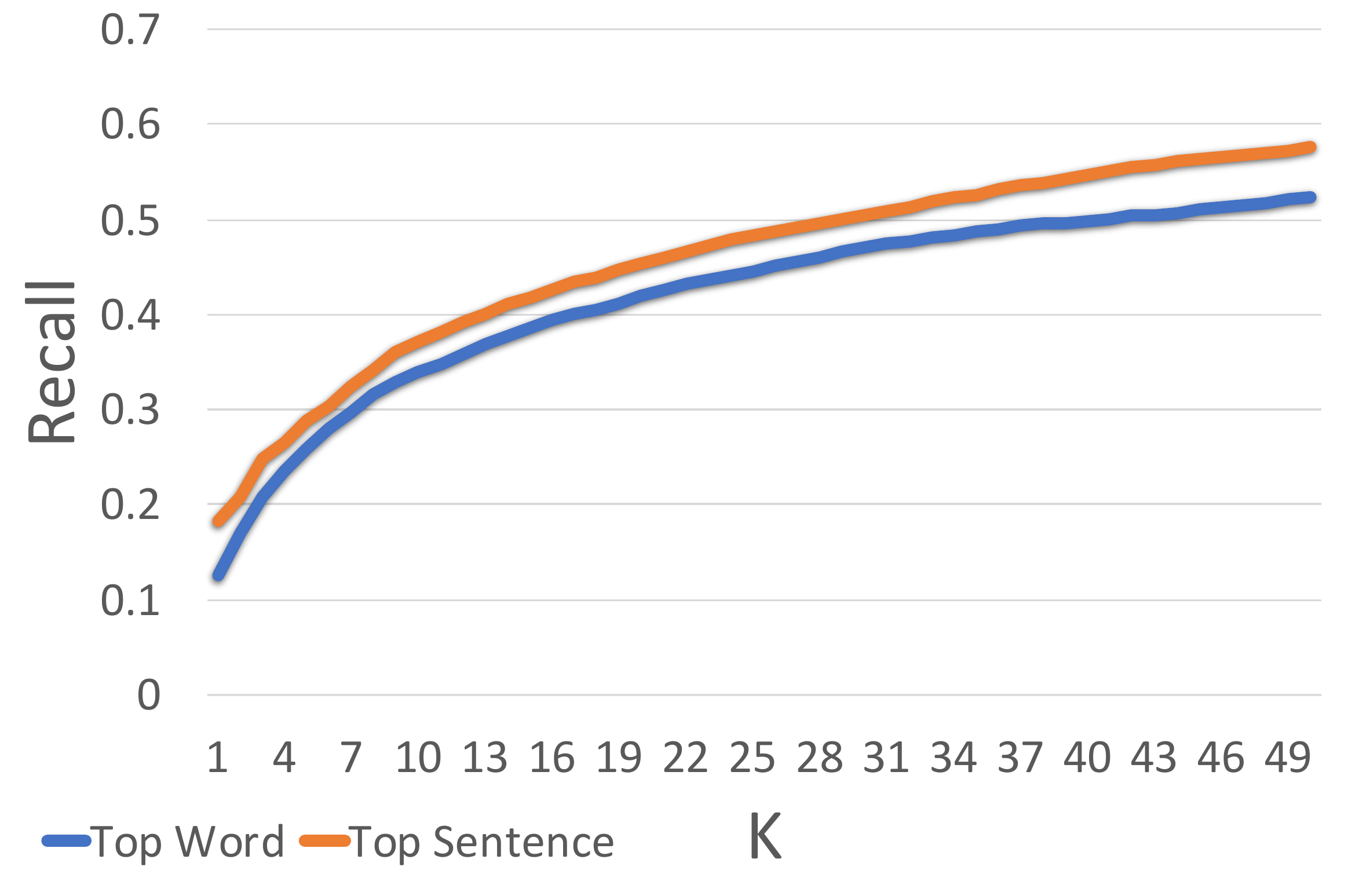}
         \caption{Retrieval@K curve for words and sentences.}
     \label{fig:topk}
     \vspace{-.3cm}
\end{figure}
\subsection{ArticleNet Performance}
\label{appendix:articlenet_perf}
We rank each sentence during evaluation by the sentence score $a_{sent}$, and then plot on average how many sentences should be retrieved to find one including the answer. We compute the same curve for words where the ranking is based on the word score $a_{w_i}$ multiplied by the sentence score $a_{sent}$. Product of these scores results in a higher retrieval than $a_{w_i}$ by itself. These results show that ArticleNet is able to retrieve relevant sentences and words from the articles with reasonable accuracy. The plots that show Recall for top $K$ sentences or words are shown in Figure~\ref{fig:topk}.
\clearpage
\clearpage
\section{MUTAN+AN and BAN+AN Details}
\label{appendix:combine_details}
We provide more details for the MUTAN+AN and BAN+AN models in this section. The MUTAN model is the Multimodal Tucker Fusion (MUTAN) model \cite{ben2017mutan}. Specifically, we use the attention version of MUTAN, and choose the parameters to match the single best performing model of \cite{ben2017mutan}.

The BAN model is the single model version of Bilinear Attention Networks~\cite{Kim2018}. We use the single model version, and we use faster-rcnn features trained on COCO train (to avoid overlap with our test set). For both BAN and MUTAN, we use the top 2000 answers in train as our answer vocabulary.

We incorporate hidden states of ArticleNet for the top retrieved sentences  into MUTAN and BAN. During VQA training and testing, we take the hidden states for the top $N_{art}$ predicted sentences (ignoring duplicate sentences), and feed them in the memory in an end-to-end memory network \cite{sukhbaatar15}.

We use the visual features $V$ and encoded question $Q$ passed through a hidden layer as the key to the memory network. To incorporate the memory network into the VQA system, we concatenate the output of the memory network to the hidden layer of the MUTAN after the attention MUTAN fusion and before the final MUTAN fusion. For the BAN model, we feed the output of the question embedding as the key to the memory network, and concatenate the output of the memory network to BAN right before the final classification layers.

\section{Training and Model Details}
\label{appendix:training_details}
\subsection{ArticleNet}
The question is encoded using a pre-trained skip-thought \cite{kiros2015skip} encoder. All fully connected layers (except at output layers) have batch normalization \cite{ioffe2015batch} and ReLU activations. All output layers have Sigmoid before the final output. We train ArticleNet for $10,000$ iterations with a batch size of $64$ using ADAM \cite{kingma2014adam} with a learning rate of $10^{-4}$, and using a balanced training set of ``positive'' and ``negative'' articles, meaning that with equal probability, an input article will contain the answer somewhere.

\subsection{VQA Models}
The MUTAN models as well as the MLP and Q-Only models were trained for $500$ epochs. All use batch size of $128$ using ADAM \cite{kingma2014adam} with learning rate $10^{-4}$.

The BAN models were trained for $200$ epochs. We found that setting $\gamma$ (number of glimpses) to 2 and the hidden feature size to $512$ yielded much better performance on our dataset than the default parameter options used for VQAv2~\cite{goyal2017making}.

We choose $N_{art}$ to be $20$, number of hops in the memory network as $2$, and the hidden size of the memory network as $300$.

\section{Additional Dataset Examples}
\label{appendix:dataset_examples}
\begin{figure*}[h]
\begin{center}
   \includegraphics[width=\linewidth]{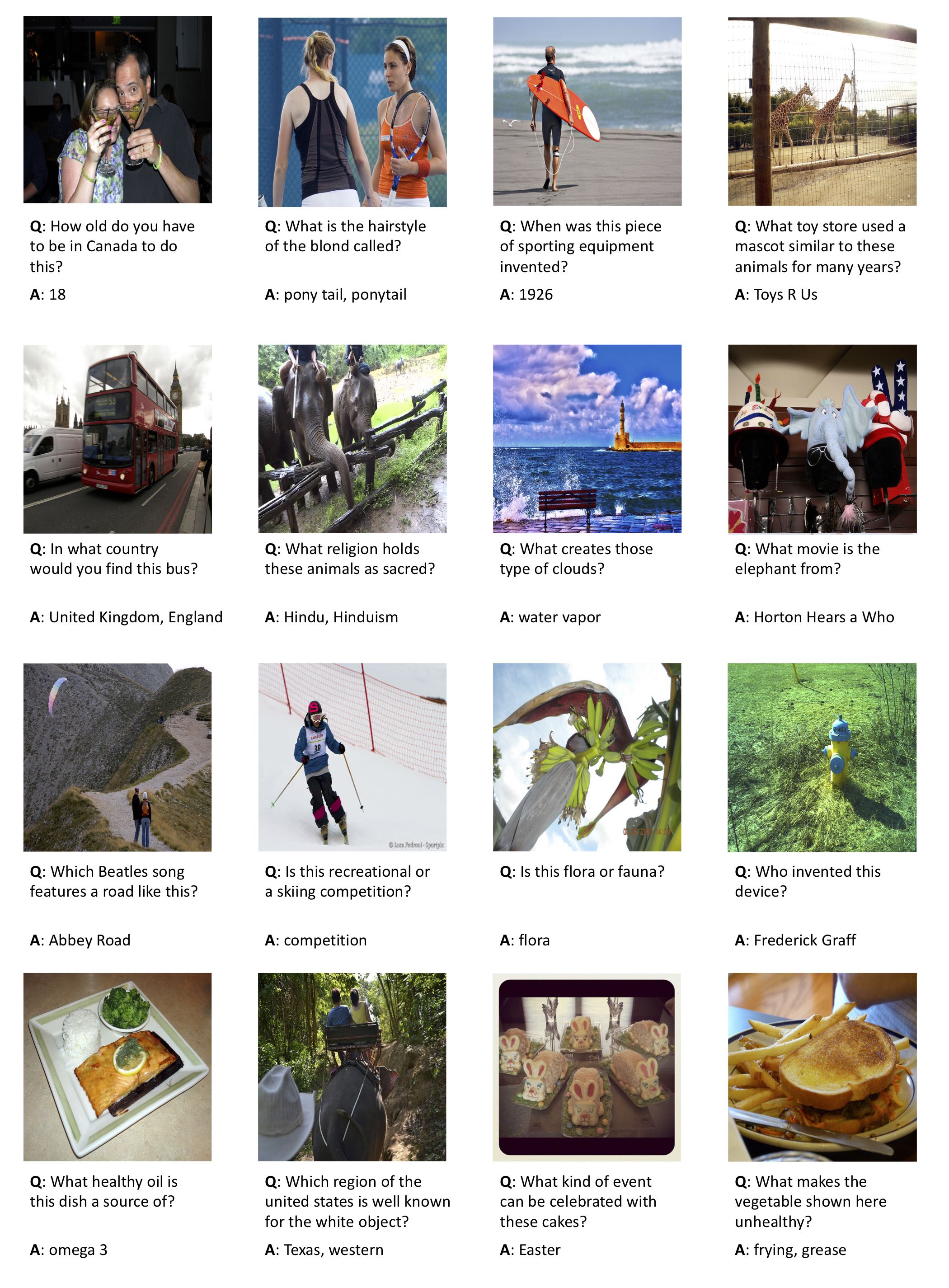}
\end{center}
   \vspace{-.2cm}
   \caption{\textbf{Dataset examples.} Some more sample questions of  \abvdataset.}
\label{fig:okvqa1}
\vspace{-.6cm}
\end{figure*}

\begin{figure*}[h]
\begin{center}
   \includegraphics[width=1\linewidth]{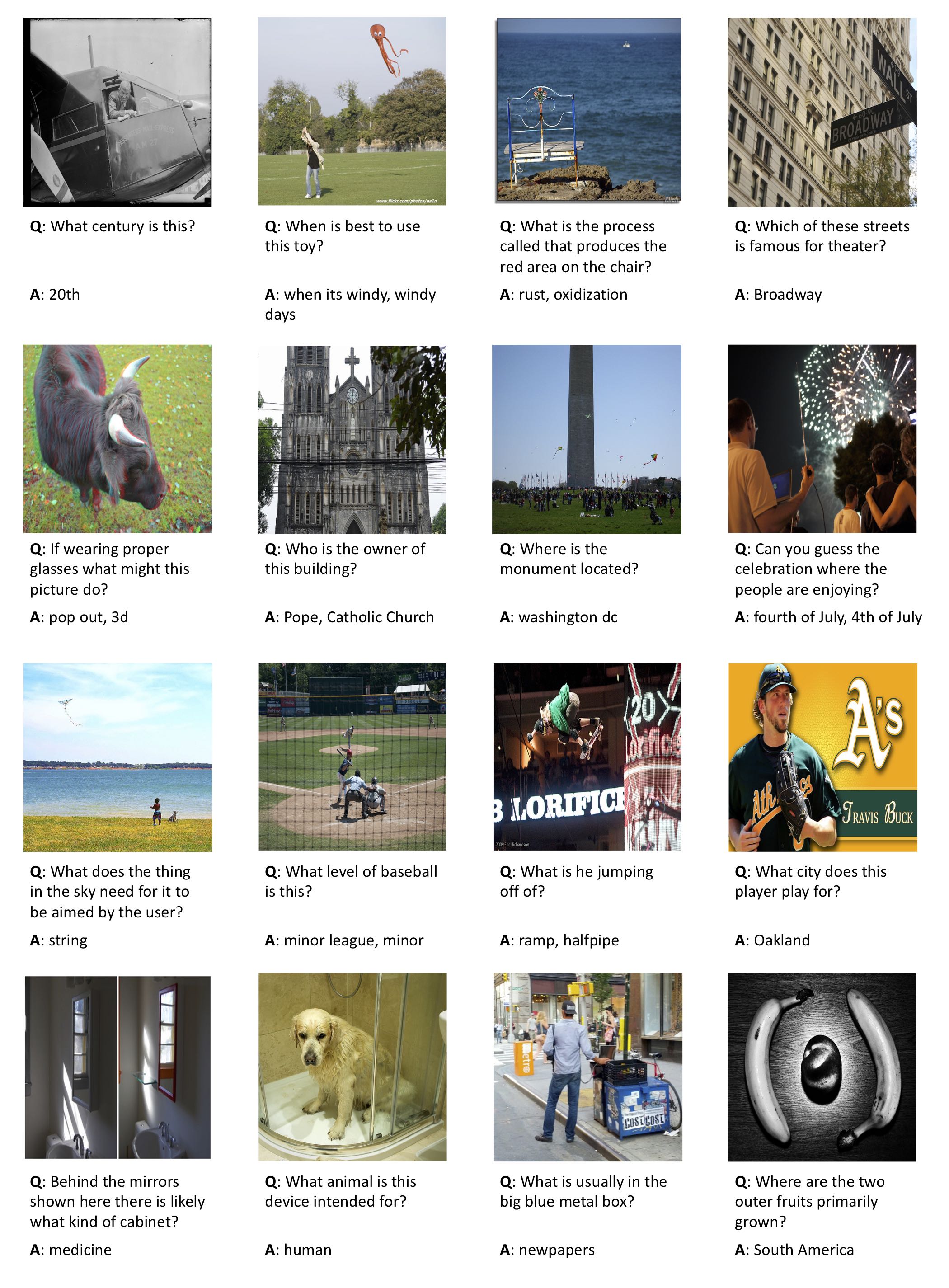}
\end{center}
   \vspace{-.2cm}
   \caption{\textbf{Dataset examples.} Some more sample questions of  \abvdataset.}
\label{fig:okvqa2}
\vspace{-.6cm}
\end{figure*}

\begin{figure*}[h]
\begin{center}
   \includegraphics[width=1\linewidth]{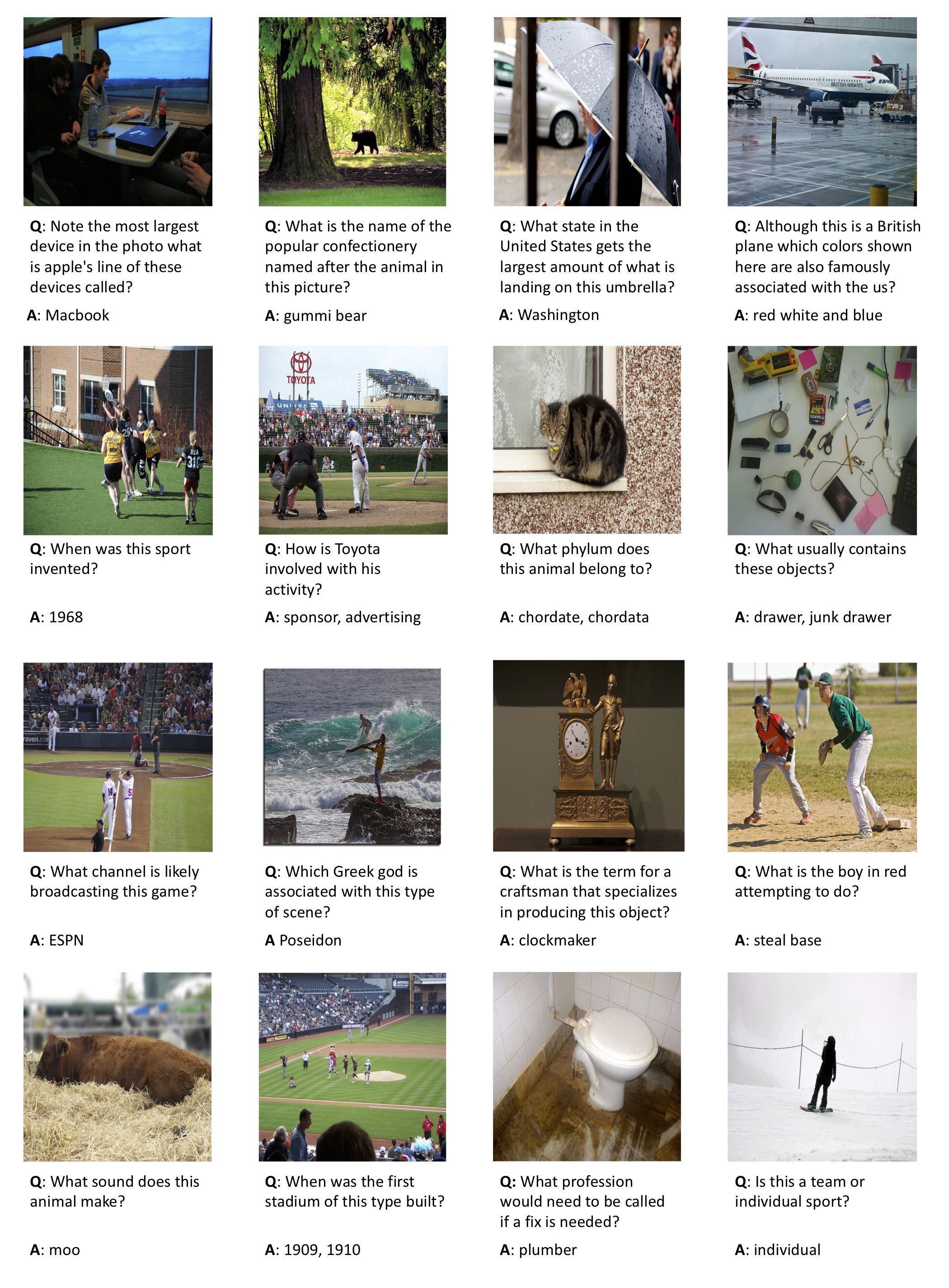}
\end{center}
   \vspace{-.2cm}
   \caption{\textbf{Dataset examples.} Some more sample questions of  \abvdataset.}
\label{fig:okvqa3}
\vspace{-.6cm}
\end{figure*}

In Figures~\ref{fig:okvqa1},~\ref{fig:okvqa2},~\ref{fig:okvqa3} we provide additional examples of \datasetname.


\end{document}